# Analyzing the Impact of Varied Window Hyper-parameters on Deep CNN for sEMG based Motion Intent Classification

Frank Kulwa, Oluwarotimi Williams Samuel*, *Senior Member, IEEE,* Mojisola Grace Asogbon, *Member, IEEE,* Olumide Olayinka Obe, and Guanglin Li*, *Senior Member, IEEE*

*Abstract—* The use of deep neural networks in electromyogram (EMG) based prostheses control provides a promising alternative to the hand-crafted features by automatically learning muscle activation patterns from the EMG signals. Meanwhile, the use of raw EMG signals as input to convolution neural networks (CNN) offers a simple, fast, and ideal scheme for effective control of prostheses. Therefore, this study investigates the relationship between window length and overlap, which may influence the generation of robust raw EMG 2-dimensional (2D) signals for application in CNN. And a rule of thumb for a proper combination of these parameters that could guarantee optimal network performance was derived. Moreover, we investigate the relationship between the CNN receptive window size and the raw EMG signal size. Experimental results show that the performance of the CNN increases with the increase in overlap within the generated signals, with the highest improvement of 9.49% accuracy and 23.33% F1-score realized when the overlap is 75% of the window length. Similarly, the network performance increases with the increase in receptive window (kernel) size. Findings from this study suggest that a combination of 75% overlap in 2D EMG signals and wider network kernels may provide ideal motor intents classification for adequate EMG-CNN based prostheses control scheme.

## I. INTRODUCTION

The loss of upper limb leads to difficulty in performing simple and complex daily activities which can cause emotional and psychological problems that decrease the quality of life of amputees. In order to restore limb functionality in amputees, researchers have focused on developing prostheses based on pattern recognition of motion intents from surface electromyogram (sEMG) signals. The sEMG signals present the sum of underlying motor action potential produced from the limb muscle contraction [1] that could serve as a potential control source for intelligent prostheses. The stochastic and non-linearity nature of sEMG have motivated the use of engineered features towards adequate characterization of hidden patterns associated with the motion intents. In that respect, Li et al. used a combination of time domain features to train a regulator-based classifier in mitigating the effects of muscle contraction force variation when decoding motion intents from sEMG signals [2]. Likewise, to resolve the coexisting impact of subject mobility and muscle contraction force variation on motion intents from sEMG, Asogbon et al. proposed five invariant time-domain features [3]. To improve the recognition of limb movements, both time and spatial characteristics of the sEMG signals, Samuel et al. introduced engineered spatio-temporal descriptors which are robust to additive Gaussian noise [4]. Similar features were applied in [5] for optimal decoding of movement intents in upper limb of the stroke survivors. While these studies considered instantaneous (raw) sEMG signals not suitable for prostheses control [6] and required feature engineering, deep learning techniques have shown that feature learning methods can be applied directly to such signals [7].

The sEMG-based upper limb gesture classification task can be modeled as a two-dimensional (2D) signal recognition problem using convolution neural networks (CNN), where the input signal has a size of K ×T× 1 (width x length x depth). Various methods have been deployed for constructing 2D signals such as; construction of 2D signal from HD-sEMG electrode array shape where the 2D depicts the array placement during data collection [8], the use of spectrograms generated from the short-time Fourier transforms of sEMG segments [9], and generation of 2D signals from chunks of sEMG signals using overlapping time windows (where the width matches the number of electrodes and the length is equal to the window size). Although recent research suggests that focusing on the temporal information of the signals can help CNN extract long and short-term patterns [8], [10], [11], they are well-suited for extraction of spatial features. With regards to sparse EMG collection configurations, the studies show that the choice on the size of the analysis window and overlap (during creation of 2D signals) has been made empirically to trade-off the classification performance and computation time [12] and there is no rule of thumb to follow. For example, [13] used window length of 150ms, [14] used 150ms window length and overlap of 90ms, [15] applies 100ms and increment of 10ms which is equivalent to 200 samples per window and overlap of 180 samples, [16] used window of 150 data points and overlap of 30 data points, and [17] applied window of 300 ms and stride of 10 ms.

Although there are some studies that have analyzed the effects of window parameters [8, 18-21], they specifically

This work was supported in part by the National Natural Science Foundation of China (#81927804, #82050410452, #62150410439), Shenzhen Governmental Basic Research Grant (#JCYJ20180507182508857), Shenzhen Institute of Artificial Intelligence and Robotics for Society, Shenzhen Governmental Collaborative Innovation Program (#SGLH20180625142402055).

F. Kulwa, O.W. Samuel, M.G. Asogbon, and G. Li are with the CAS Key Laboratory of Human-Machine Intelligence-Synergy Systems, Shenzhen Institute of Advanced Technology (SIAT), Chinese Academy of Sciences (CAS), Shenzhen, Guangdong 518055, China. (Correspondence: Dr. Oluwarotimi Williams Samuel, e-mail: samuel@siat.ac.cn and Dr. Guanglin Li, e-mail: gl.li@siat.ac.cn).

F. Kulwa is also with the Shenzhen College of Advanced Technology, University of Chinese Academy of Sciences, Shenzhen, Guangdong 518055, China.

O.O. Obe is with the Department of Computer Science, Federal University of Technology, Akure, Nigeria.

focused on the extraction of engineered features and only utilized conventional machine learning methods without considering deep neural networks. To the best of the authors' knowledge, no study has analyzed the effects of window parameters (window length and overlap) on 2D EMG signals for deep neural networks particularly CNNs to date. Thus, this constitutes a research gap that the current study seeks to address. Hence, this paper investigates the relationship between window parameters (such as window size and increment) on the generation of 2D raw EMG signals towards yielding optimal classification results for the deep neural networks. Moreover, we examine the impacts of receptive window sizes (kernels) for varied configurations of CNNs and how they impact the feature learning of sEMG signal.

## II. METHODS

### A. Subjects

A total of four transhumeral amputee subjects took part in the sEMG data collection experiment in this research. Their ages ranged from 35 to 49, and two of them were dominant in their right hand. The subjects were briefed about the study's goal and purpose prior to data collection, and they all agreed to participate. Following that, the subjects signed a written informed consent form and agreed to their data being published for scientific and educational purposes. The experimental protocol of the study was approved by the Institutional Review Board of the Shenzhen Institutes of Advanced Technology, Chinese Academy of Sciences, China.

### B. Equipment setup and data acquisition

Five motion classes of wrist supination (WS), hand open (HO), wrist pronation (WP), hand close (HC), and no movement (NM) were investigated. A computer screen was set up in front of the participants during the experiment. A motion picture would be shown for each of the motion classes engaged in the study. When a picture of the target movement is shown on the screen, the subjects perform it. And when it disappeared, the subject stopped doing the movement. The five motion classes were performed at random with a comfortable force level decided by the participants to avoid muscular and mental tiredness.

Each motion lasted 5 seconds, with a 5-second rest period (NM) in between two adjacent motions. Each subject had five data recording sessions, each of which comprised of 40 active motions (each of HO, HC, WP, and WS appeared 10 times at random) and 40 repetitions of NM. A high-density sEMG system (REFA 128, TMS International, the Netherlands) was used to acquire the sEMG signals, with 32 monopolar electrodes implanted on the skin surface of the residual arm for each amputee.

### C. Data preprocessing

The sEMG data corresponding to each of the indicated classes of limb movement was preprocessed and evaluated using the MATLAB programming tool. To be more specific, the data was sampled at a rate of 1024Hz, with a 50Hz notch filter used to reduce power-line interference. Variations in EMG signals caused by motion artifacts and electrode displacement were reduced using a band pass filter with cut-off frequencies of 10 and 500 Hz.

### D. Generation of 2D signal

In order to simplify the network's learning process during training, the data was first normalized by Z-score normalization technique with unit standard deviation and zero mean. During pilot experiment, Z-score and Mean-Max normalization methods were evaluated on two types of dataset configurations. Firstly, the dataset was normalized across a particular class of movement. That is, the data corresponding to HC movement of a particular subject is normalized as a whole. Secondly, the normalization was done segment-wise (2D signals). Finally, we considered using Z-score on class-wise normalization due to the stable and high results obtained for this method.

Then the 2D signals were generated using segmentation window of size $K \times T \times 1$, where K is the number of electrodes which is 32 for this study and T is the window length. To investigate the effects of window lengths and overlap on the robustness of the generated 2D signals, three different window lengths were examined, window length 125ms (data points), 150ms and 175ms. These values were selected because they have been suggested to give optimal results and used by number of studies [20]. To set a rule of thumb we vary the overlaps in percentage of the window length. Therefore, for each window length the overlaps are varied at 75% of T, 50%, 25% and 0% of T (No overlap).

### E. Convolutional neural network (CNN) architecture

We propose a model which consists of four feature learning blocks, feature reduction layer and one feature selection layer as shown in Figure 1.

The first block consists of convolution layer with 32 nodes, ReLu activation function which is followed by 10% dropout layer so as to reduce the overfitting and improve generalization [22].

To allow kernel to cover more space within the feature maps we keep the spatial dimension of the signal features the same before and after each convolution by applying padding.

The other three blocks have similar layers packing with only difference in the number of nodes, which are 32, 64 and 64, respectively. Each block consists of a convolution layer followed by ReLu and dropout of 10%. To make the network more robust to changes of the input signals such as translation and shifts, we apply max pooling layer of size 2x2 at the end of each block. After feature learning by the convolution blocks, the feature maps are passed through global maxpooling for dimension reduction. We use global max maxpooling because it is not prone to over-fitting and improves the model generalization.

The bottom part of the network consists of a dense layer of 128 nodes for more feature selection that is relevant to classes. Lastly, the classification of the 5 limb motion intents using a softmax layer.

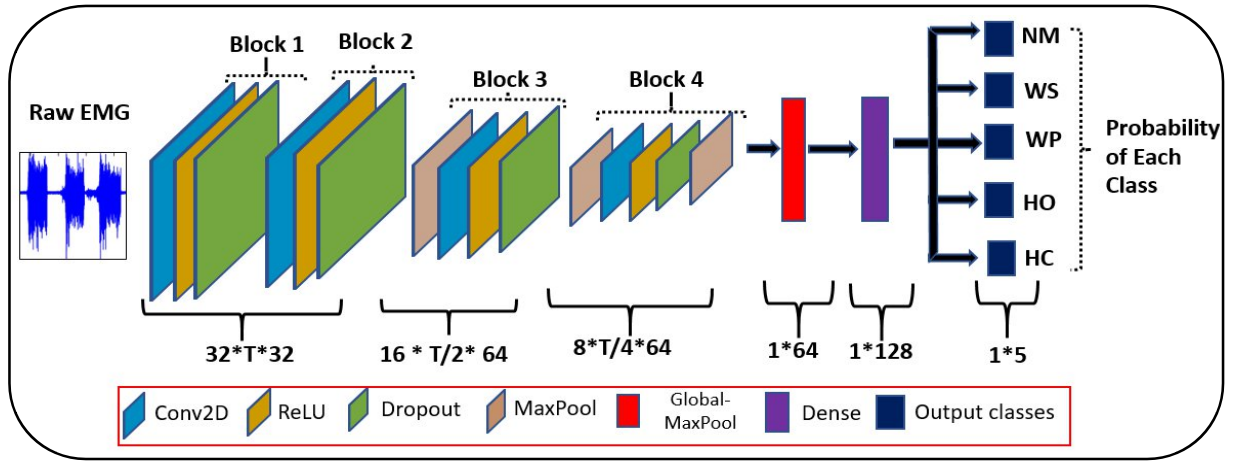

Figure 1. The architecture of the proposed CNN model. Where the output size of Block 1 and 2 is 32*T*32, Block 3 is 16*T/2 * 64, and Block 4 is 8*T/4 *64. T represents the window length. The output classes at the final layer are no movement (NM), wrist supination (WS), wrist pronation (WP), hand open (HO) and hand close (HC).

To investigate the impacts of receptive window sizes (kernels) of the network while learning general or fine features within the EMG signal, we have designed three networks of similar architecture but with different kernel sizes. For each network the same size of convolution kernel is applied in all layers. We use 3x3, 5x5, and 7x7 sizes because they are the fundamental kernel sizes which have been applied in many studies and give optimal network performances [9, 13–15].

*F. Experimental setup*

All the models were trained using the Adam optimizer with the learning rate of 0.0001 as it showed optimal network performance during our preliminary experiments compared to SGD optimizer. Moreover, we used a categorical cross entropy as the loss function due to its better performance in multiclass problems [23].

Observing the training and validation curves of the pilot experiment in Figure 2, the curves tend to flatten at 30 epochs. Thus, we used 35 epochs in all training experiments to allow more confident results. The outcomes of all models are presented using the fundamental metrics for classification problems, accuracy and F1-score.

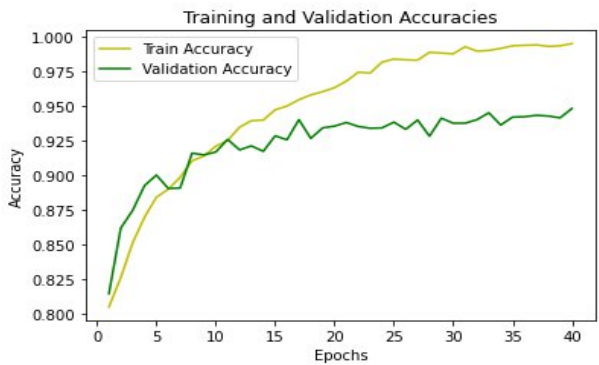

Figure 2: Training and validation curves to test the possible number of epochs

### III. RESULTS

*A. Evaluating the impact of window overlap and size on the robustness of the generated 2D EMG signal.*

To analyze the effect of data overlap with respect to segmentation (analysis) window, the window length (T) was kept constant while varying the overlap per experiment. Then, the generated 2D signals for each variation were used to train the three different CNNs (with kernels sizes of 3, 5, and 7) across each subject's data. As suggested by previous studies [16], 70% of the data was used for training the networks and the remaining 30% for testing. The overall average classification results obtained across all subjects were shown in Figure 3. Besides, we only presented results for kernel sizes of 3 and 5.

By carefully observing the results in Figure 3, it can be seen that the classification performance increases with the increase in percentage of overlap regardless of the 2D signal size (window length (T)) in all proposed deep neural networks. For instance, in Figure 3(b) when the network with kernel =5 and window length of 150 is used, the accuracy of 97.80% is achieved at the overlap of 0.75T and 89.04% accuracy at the overlap of 0T. Similarly, F1-score of 94.46% and 72.80% are observed at overlaps of 0.75T and 0T, respectively. This indicates an increment of approximately 8.76% accuracy and 21.66% F1-score. Meanwhile, the highest improvement of 9.49% accuracy and 23.33% F1-score is achieved when kernel=5 and T=175 in Figure 3(c). The results above show that the 2D signals with an overlap of 75% can achieve acceptable results for classification of upper limb motion intents based on sEMG signals regardless of the window length.

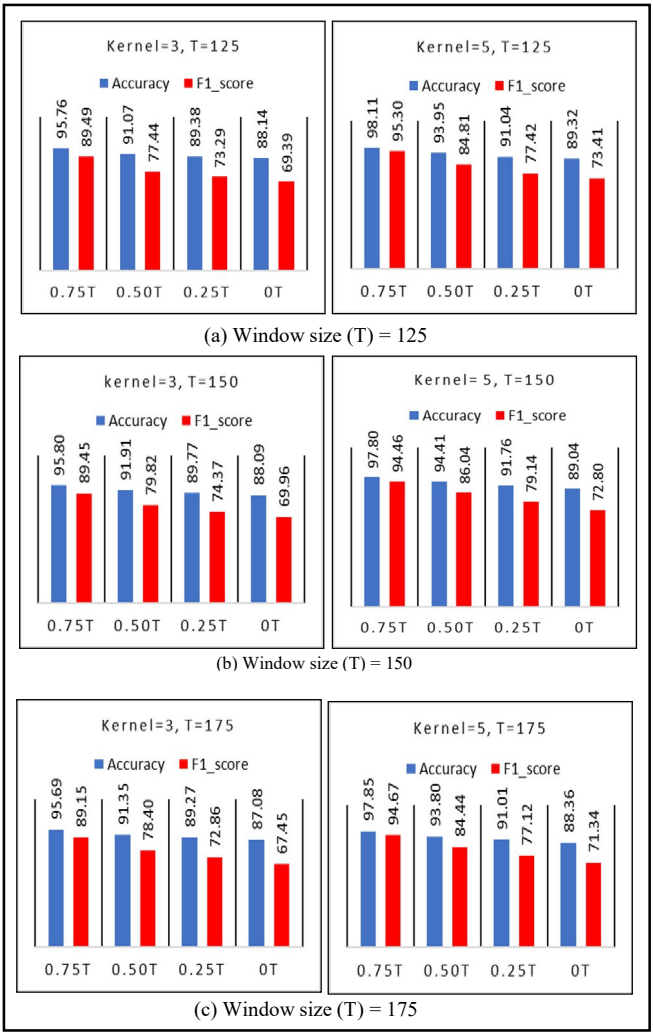

Figure 3. Limb motion intent classification accuracy (Blue bars) and F1-score (Red bars) across two networks (kernel=3, 5) at window size (T), the overlap is applied at 75% of T (0.75T), 50% (0.5T), 25% (0.25T) and No overlap (0T). (a) When the 2D signals are of size 32x125 (b) the signals are of size 32x150 and (c) when the size is 32 x 175.

### B. Evaluating the effects of network receptive fields against the generated 2D EMG signal sizes

In this section the performance of the proposed networks (kernel of 3, 5, and 7) are examined on different dimensions of the 2D signals (at window length (T) of 125, 150, and 175). Because of the relatively higher results obtained using signals with overlap of 75%, only this overlap was considered. Thus, the classification results based on F1-score metric for the three CNNs at different window sizes are presented in Figure 5.

Observing the results in Figure 5, it can be seen that the increase in kernel size led to increase in performance of the networks. For example, at signal window length of 125 the three networks with corresponding kernels of 3, 5, and 7 achieved 89.49%, 95.30% and 95.93% F1-score, respectively. And this resulted to an increase of about 6.44% in F1-score between kernel 3 and 7. Similarly, an increment of 5.40% F1-score is observed when T=150, and the highest increment of 7.79% F1-score is attained when the window size is 175. Although high impact is seen in the variation of kernel sizes, there is no much effect when varying the signal window length on the performance of the networks.

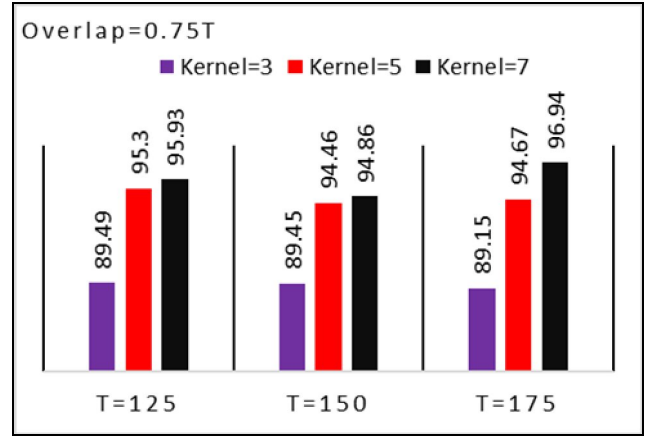

Figure 5: The classification results of different networks (at kernels=3,4,5) with variation in 2D window length size (T=125, 150, and 175). The results are in percentages F1-score.

### C. Analysis of individual motion intention across window overlaps and kernel sizes

The analysis in the previous sections only present rather general view of window sizes, overlaps, and kernel sizes impact on motion intent classification. Therefore, there is need to examine the impact of these parameters on the decoding performance of individual class of movement. Thus, the effect of kernel size and overlap on the characterization of individual class of motion is presented in Figure 6. The effects of kernel sizes and signal window overlap were investigated on each MI. It is noteworthy that only the highest performing kernel (kernel=7) and the lowest (kernel=3), likewise for the overlap windows (0.75T and 0T) across individual class of motion were considered as shown in the confusion matrices in Figure 6. This is because the window length only exhibits slight impact on the classification performance. Thus, we used T=175 for representation only and the confusion matrices are presented as an average across all subjects.

From the classification results indicated in Figure 6, it can be seen that the lowest accuracy is shown for Hand Open (HO) motion intention by both networks (*kernel equal 7 and 3*) when there is no overlap, and kernel =3 has lower results compared to kernel=7. Classification accuracy higher that 95% was recorded for all motion intentions when the kernel=7 and the overlap is 0.75T as shown in Figure 6 (c). This shows that the network of kernel =7 and 2D signals when overlap is 75% would be an ideal combination for characterizing the EMG signal patterns related to these classes of hand movements compared to others.

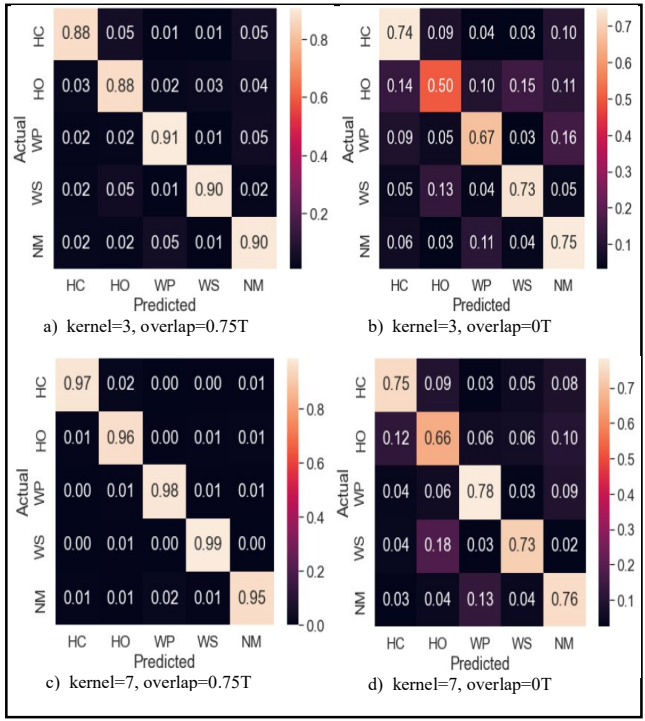

Figure 6: Classification performance of kernel 3 and 7 when the overlap is 75% and 0% of window length (T) across each motion intent.

IV. DISCUSSION

Feature learning through the use of deep neural networks represent a promising alternative to hand crafted features which consume a lot of time for preparation. To achieve clinical use of DNN based prostheses, a relatively robust and simple signal processing scheme (which is independent of the structure of the classification network) is needed.

Therefore, in this study we explore the use of raw sEMG signals in the identification of motor intents based on convolution neural networks. Moreover, we investigate the relationship between the 2D signal length and overlap that can give optimal performance of the network. Additionally, we examine the effects of kernel sizes of the CNNs with respect to classification performance of the model when applying raw EMG signals.

From the analyzed results of Figure 3, the percentage of overlap within the generated 2D signal highly contributes to the performance of the CNNs. The trend shows that 75% of overlap yielded the highest performance for the networks with an improvement of 9.49% in accuracy and 23.33% in F1-score compared to without overlap. And the classification performance increases with the increase in overlap. From this, we can see that a positive correlation exists between the *performance of deep neural network* and *percentage of data overlap*, leading to the following rule of thumb: ***The classification performance of Convolutional neural networks increases with increase in the percentage of data overlap at a constant window length.***

Due to the non-stationary nature of the EMG signal, it is possible to have 2D signals segmented at different time window from the same class and subject to have different probabilistic distributions [24]. This can be clearly observed when the signals are generated without any overlap. The lack of common distribution of data from the same class hinders the optimal performance of the classification network [25]. The study presented in this paper show that a common distribution between samples (segments) from the same class can be achieved by increasing the percentage of overlap which allows the model to learn easily the hidden distribution (patterns) within the data and increase the performance. This can be justified more by the analysis presented in Figure 6 where the classification results across individual motion class show that 75% signal overlap has higher outcomes in all five classes compared to when there is no overlap.

The kernel of the network can be considered as an eye of the network, the wider the kernel, the wider the receptive field of the network, meaning that it can have a broader range of view over the signal and learn more general spatial features (patterns) while a small kernel size learns fine features. [15]. The results obtained in Figure 5, show that the increase in kernel size increases the performance by a considerable margin regardless of the window size, with the highest performance obtained by kernel 7. This implied that the pattern of the EMG signals is present over a large spatial dimension of the signal [26]. Moreover, a wide field of view allows the network to model hidden temporal connectivity within the signal [15]. Thus, it can learn both spatial and temporal trend (patterns) of the signal.

It should be noted that in this study we focused only on variation of three window lengths and kernel sizes which are among the commonly used values. Since some studies that focused on traditional machine learning and feature engineering recommended that the window length can be as lower as 32 and as higher as 300ms, in the future work, further analysis will be carried out on these extreme ranges and more kernel sizes will be considered with various CNNs and other deep neural network architectures. Moreover, we will investigate the robustness of the 2D signals at higher percentage overlap in the presence of Gaussian noise which are sometimes inherent in practical use of the prostheses.